# An integrated Auto Encoder-Block Switching defense approach to prevent adversarial attacks

Ashutosh Upadhyay, Anirudh Yadav, S.Sharanya

*Abstract* — According to the recent studies, the vulnerability of state of the art Neural Networks to adversarial input samples has increased drastically. Neural network is an intermediate path or technique by which a computer learns to perform tasks using Machine learning algorithms. Machine Learning and Artificial Intelligence model has become fundamental aspect of life, such as self-driving cars [1], smart home devices, so any vulnerability is a significant concern. The smallest input deviations can fool these extremely literal systems and deceive their users as well as administrator into precarious situations. This article proposes a defense algorithm which utilizes the combination of an auto-encoder [3] and block-switching architecture. Auto-coder is intended to remove any perturbations found in input images whereas block switching method is used to make it more robust against White-box attack. Attack is planned using FGSM [9] model, and the subsequent counter-attack by the proposed architecture will take place thereby demonstrating the feasibility and security delivered by the algorithm.

## I. INTRODUCTION

Machine Learning (ML) has spread its wings in almost all domains from medical to industrial equipment maintenance [1]. Adversarial examples can be interpreted as optical illusions to ML models in layman terms. These examples are carefully perturbed as inputs to the ML models which subsequently generate erroneous outputs. While such perturbations may seem benign to human perception, it can elicit wrong predictions from the model with full confidence. For example, perpetrators could target self-driving vehicles by causing perturbations using paint or stickers that will cause vehicle to decipher the sign incorrectly. Adversarial examples depict that many modern ML algorithms can be deluded in incredibly simple ways.

Such failures indicate that even simple ML models can have their behaviour manipulated. Traditional mechanisms for building robust ML models generally do not provide a pragmatic defence against adversarial examples. There are two effective methods that have provided a somewhat effective defence: adversarial training and defensive distillation. This article explores a much more superior defence algorithm against such adversarial examples, thereby eliminating the possibility of white-box attacks.

### 1.1 Definitions

**Adversarial Training:** It is a brute force solution which improves the model's robustness by incorporating adversarial examples into the training stage followed by an explicit training of the model against such deceptions. In simple terms, it is the process of creating and incorporating adversarial examples into the training segment.

**Defensive distillation:** The ML model is trained to output probabilities instead of making hard decision of classifying input into different classes. It is a defensive strategy.

**White-Box attack:** It is the type of attack in which attackers have access to the Machine Learning algorithm or model parameters, hyper-parameters, architecture, weights, to name a few.

**Black-Box attack:** In these type of attacks, attackers do not have any kind of information regarding the Machine Learning algorithm/model.

The following networks are used in the proposed architecture:

### 1.2 Fast Gradient Sign Method

Fast Gradient Sign Method (FGSM) [9] is the powerful attack method. It is used to attack on neural network by the means of gradient. FGSM utilises the back propagated gradients of loss with respect to image input. In other words, instead of dynamically modifying weights based on back propagated gradients, FGSM manipulates input image to maximize the loss with respect to same back propagated gradient.

$$adv\_x = x + \epsilon * \text{sign}(\nabla_x J(\theta, x, y))$$

Where-
- adv_x: Adversarial Image
- x: Original Image
- y: Original input label
- ε: Multiplier to ensure the perturbations are small.
- Θ: Model parameters
- J: Loss

### 1.3 Generative Adversarial Networks

Generative Adversarial Networks (GAN) [6] uses unsupervised learning to figure out patterns and irregularities in the input data and also use them to generate new output that has the same patterns and irregularities as the input data. The architecture is shown in fig 1.

**1.3.1 Generator:** It takes a random set of input values of fixed length, and it tries to generate a sample in the domain same as of input data. Input is drawn randomly from a Gaussian distribution. Once the model is trained, it can be used in future for generating new samples as same as input data.

**1.3.2 Discriminator:** It is a classification model which classifies the input data into real and fake classes. Here, real stands for an original data set and fake stands for data generated by generator module. It is only used for training purpose as it assists the generator in generating output samples as similar to input samples. Once training is completed discriminator is discarded.

These two models work together as discriminator provides feedback on the fake generated inputs and generator then uses these to improve output. These two are run until the discriminator model is in such a state that it can no longer distinguish between fake and real data set.

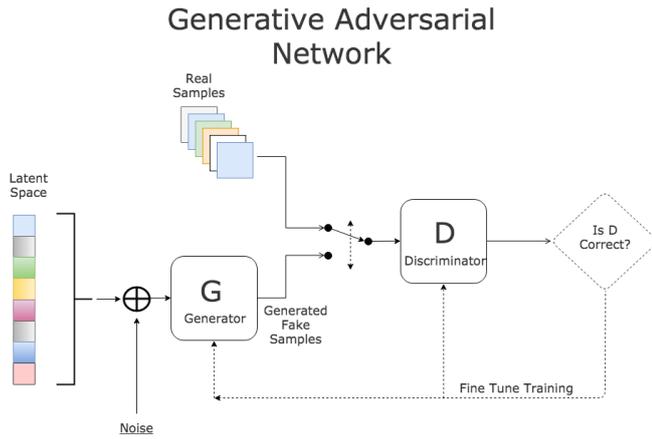

Fig 1. Architecture of Generative Adversarial Networks.

**AdvGAN**

AdvGAN (Adversarial GAN) is a modified version of GAN. However, instead of generating samples the same as of input data, it tries to generate samples that can fool state of the art models. The working of Adversarial GAN is similar to GAN but it consists of 3 models:

**Classifier:** It checks whether the output generated belong to the desired class (different from its correct class). It ensures that the generated sample gets incorrectly classified. Since discriminator ensures that samples generated by the generator are looking similar to the input data and classifier ensures that it gets wrongly classifies, so a similar-looking output to a human eye easily fools state of the art neural networks.

**Auto-Encoder:** These are Neural Networks that used Unsupervised Learning to perform data compression, denoising, dimensionality reduction, etc. The layout of auto encoders are given in Fig 2. It consists of 3 parts:

**Encoder:** This part of the auto-encoder compresses the input data into a latent space. Latent space is a compressed representation of the input data in a reduced dimension.

**Code**: This part of auto-encoder is also known as a bottleneck. It represents compressed input in a reduced dimension and passes it to the decoder.

**Decoder:** It takes latent space representation (compressed input) as an input and tries to generate the output of the same dimension as of original input. This reconstruction is done depending on the purpose. It can be used to remove noise from an image or can be used for data compression.

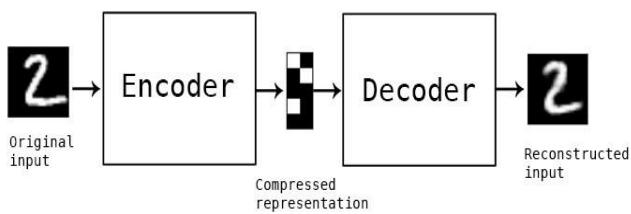

Fig 2. Auto-Encoder architecture. The input image is encoded to a compressed Representation and then decoded.

## II. LITERATURE REVIEW

| Title | Merit | Demerit |
|---|---|---|
| One Pixel attack for fooling Deep Neural Networks 2017 [2] | Requires less adversarial information | For high definition image required multiple pixel perturbations |
| PuVAE: A Variational Auto-encoder to Purify Adversarial Examples 2019 [3] | Takes only 0.114 second to process | Only filter Limited examples |
| Defensive Dropout For Hardening Deep Neural Networks under Adversarial Attacks- 2019 | Decreased the white box attack rate from 100 percent to 13 percent | Unable to defend all types of adversarial attacks |

### 2.1 COMPARITIVE STUDY

| Existing Methods | Advantage | Disadvantage |
|---|---|---|
| Anomaly Detection Model | Detects anomaly and avoid input reaching to the actual model. | Requires additional Time. It can also be termed as bottleneck for fast models |
| Pruning Method | Prevents Noise Based Perturbations. | Not efficient for Black box attacks |
| Distributed Colour Channels | Blue Channel Performs exceptionally well to determine any adversarial attacks. | Multiple model Occupy large time and space while running. Not good for video input. |
| Adversarial Dataset | Makes model robust against those Adversarial input which is trained. | Difficult to train for all adversarial attack types as it Largely affects the accuracy. |
| Random Weight switching | Good for White box attacks | Not efficient for Black box attacks |

## III. PROPOSED WORK

The proposed system majorly focuses on static image input and defence architecture. Following are the characteristics of the proposed model:

- Combination of two models to effectively defend both Black box and White Box attack.
- Randomization [8] acts as a backup for filtration performed by auto-encoder there by increasing the robustness of the proposed model.
- Grad-CAM [7] allows the model to predict the highlighted important region based on classification.

### 3.1 Defence Architecture

The defence architecture comprises of 3 major components- Auto-encoder [3], Block switching [8] and Grad-CAM [7]. The purpose of auto-encoder is to perform filtration. Auto-encoder is integrated with the Block switching model along with the Grad-CAM. Block switching consists of multiple processes which are selected randomly based on the classification of images. Grad-CAM are the activation maps which uses the gradients to produce highlights on important region in the image.

*Fig 3. Defense architecture using the combination of auto-encoder and block-switching method.*

## IV. IMPLEMENTATION

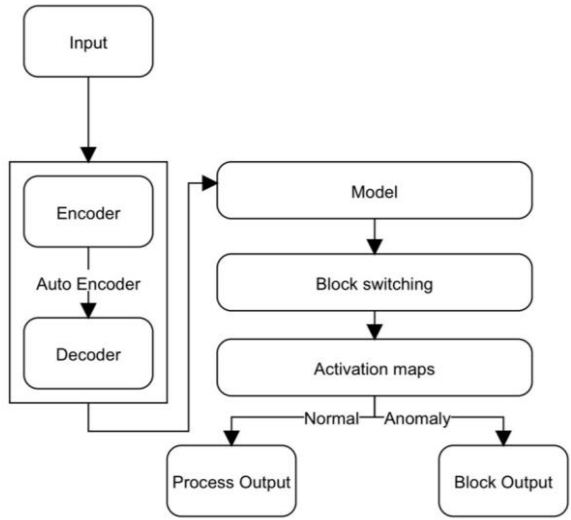

### 4.1 Auto Encoder

Auto-encoders can be used for filtration purpose by taking adversarial data in the form of inputs and clean data as outputs. It is possible for them to remove adversarial noise from an input image. Thus in the proposed architecture, auto-encoders can also be termed as denoisers.

### 4.2 Block Switching and Random Weight Switching

Block-switching [8] model is trained on two different levels. The first level of training comprises of individual training of sub-models with same architecture. Due to the same architecture of sub-models, every sub-model possess similar characteristics. Block-switching consists of multiple channels. Every sub-model is suitably divided in its lower body and upper body. Lower body contains of all the convolutional layer. Lower parts are again merged to form a single output by including parallel channels of block-switching while the other parts are discarded. Every sub-model has different model parameters due to random initialization and stochasticity in the training process, therefore they tend to have similar characteristics in terms of classification, accuracy and robustness.

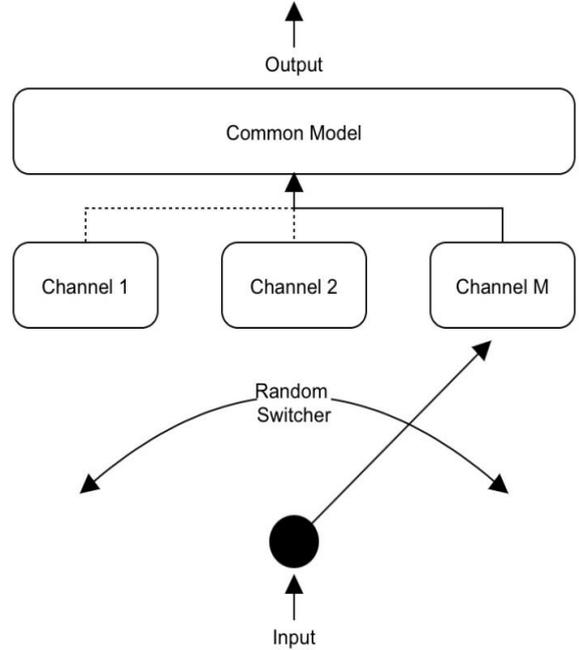

*Fig 4. Levels of Block-switching model training, and preview of multiple channels and sub-models.*

### 4.3 Grad-CAM

The aim of Grad-CAM [7] is to provide a visual explanation for model decision by using gradients flowing into the final convolution layer. Further it is used to produce a map highlighting the important region in the map.

### 4.4 Methodology

The flowchart (*Fig 5.*) depicts the entire flow of processes which are deployed in the proposed architecture. The unique combination of Auto-encoder, Block-switching [8] and grad-CAM [7] increases the accuracy and robustness of the model.

The proposed architecture follows several phases namely - image classification, building of thoroughly supervised dataset of adversarial images, auto-encoder operating as a noise remover along with block-switching containing multiple sub-channels, producing highlighted maps as an output from grad-CAM and ultimately the output of the model is classified image with highlighted important regions on it.

As an initial step in the architecture, image is taken as input for two purposes specifically – classification and building of thoroughly supervised dataset of adversarial image. Image classification is performed to ascertain the entire image. The dataset which consists of two differently classified images (original and adversarial) is provided as an input to auto-encoder. Auto-encoder removes noise from the input image and yields a denoised image as an output. The output of auto-encoder is taken as an input by block-switching model. Block-switching contains different sub-channels which are selected randomly to render the output. Grad-CAM are activation maps which generate highlights on the classified image to uncover important regions in it. The output of block-switching is taken as an input for Grad-CAM. It then processes the image obtained from the image classification model to provide a visual explanation for model decision by using gradients flowing into the final convolution layer. The output of Grad-CAM is the highlighted image with important regions in it. The output of Grad-CAM can be used for anomaly detection. It also tends to notice the attack which encompasses the above deployed tightly coupled architecture of auto-encoder and block-switching model. Ultimately the output of the above explained architecture is the classification result backed by the Grad-CAM. The proposed architecture is tightly coupled and secured in terms of dataflow and levels of classification deployed to prevent the adversarial attacks. The combination of distinct phases of the model is unique and present an overall efficient and architecture.

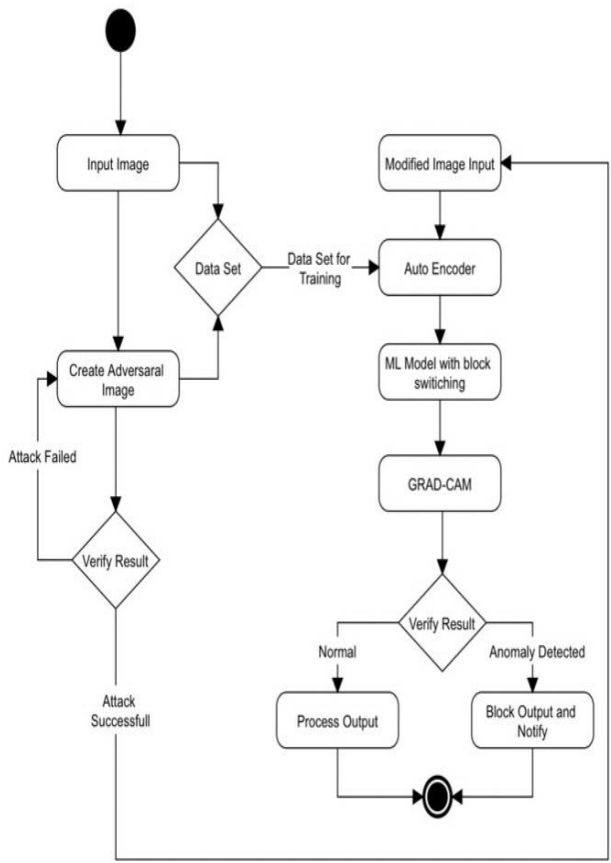

*Fig 5. Combination of auto-encoder, block-switching and grad-CAM*

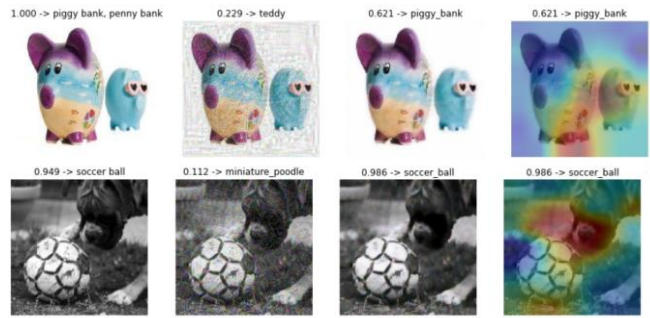

*Fig 6. Output of respective modules.*

## V. RESULTS AND DISCUSSION

### 5.1 Auto-encoder

**Input-** Attack image with perturbations

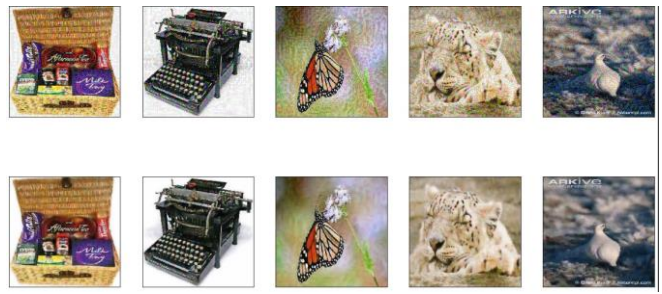

**Output-** Image is formed by removing perturbations by auto-encoder.

### 5.2 Grad-CAM

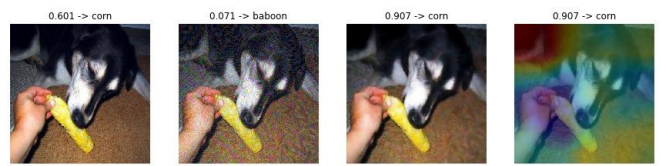

Further the image classification and grad-CAM output can be used for verification.

### 5.3 Overall Accuracy

```
-+------------------------------------------+-
| Attack Succesfully Performed in : 87 images|
| Attack failed in : 9 images                |
-+------------------------------------------+-
| Defense succesfull in : 85 images          |
| Defense failed in : 11 images              |
-+------------------------------------------+-
| Attack Model Accuracy : 87.00%             |
| Defense Model Accuracy : 88.54%            |
-+------------------------------------------+-
```

## VI. CONCLUSION

The attack by the FGSM [9] model is effectively countered by the proposed defence architecture with an accuracy of 88.54%. The unique combination of auto-encoder along with randomization in the classification model ensures efficiency, high accuracy and robustness. This work can be extended to capture images form motion videos.